\documentclass{article}

\usepackage[final]{neurips_2019}

\usepackage[utf8]{inputenc} 
\usepackage[T1]{fontenc}    
\usepackage{hyperref}       
\usepackage{url}            
\usepackage{booktabs}       
\usepackage{amsfonts}       
\usepackage{nicefrac}       
\usepackage{microtype}      

\usepackage{latexsym}
\usepackage{amsmath,amssymb,bm,color,booktabs}
\usepackage{graphicx}

\usepackage{times}
\usepackage{latexsym}
\usepackage{url}
\usepackage{graphicx}
\usepackage{booktabs}
\usepackage{array}
\usepackage{longtable}
\usepackage{multirow}
\usepackage{threeparttable}
\usepackage[T1]{fontenc}
\usepackage[utf8]{inputenc}
\usepackage[english]{babel}

\usepackage{newunicodechar}
\newunicodechar{ȟ}{\v{h}}

\newcommand*{\GtrSim}{\smallrel\gtrsim}
\makeatletter
\newcommand*{\smallrel}[2][1.0]{%
  \mathrel{\mathpalette{\smallrel@{#1}}{#2}}%
}
\newcommand*{\smallrel@}[3]{%
  \sbox0{$#2\vcenter{}$}%
  \dimen@=\ht0 %
  \raise\dimen@\hbox{%
    \scalebox{#1}{%
      \raise-\dimen@\hbox{$#2#3\m@th$}%
    }%
  }%
}
\makeatother

\usepackage{latexsym}
\usepackage{amsmath,amssymb,bm,color,booktabs}
\usepackage{todonotes}

\usepackage{tabularx}

\newcommand{\ours}{\textsc{SQLova}}
\newcommand{\modelBtoScal}{\textsc{Shallow-Layer}}
\newcommand{\modelBtoSeq}{\textsc{Decoder-Layer}}
\newcommand{\modelBtoM}{\textsc{NL2SQL Layer}}
\newcommand{\modelBaseline}{\textsc{BERT-to-Sequence}}
\newcommand{\modelBaselineT}{\textsc{BERT-to-Transformer}}

\newcommand{\oururl}{\texttt{https://github.com/naver/sqlova}}

\newcommand{\bea}[1]{ \begin{equation}\begin{aligned} #1 \end{aligned} \end{equation} }
\newcommand{\pave}[1]{\left( #1 \right)}
\newcommand{\softmax}{\text{softmax}}
\newcommand{\mcW}{\mathcal{W}}

\graphicspath{{./figure/}}

\title{A Comprehensive Exploration on WikiSQL with Table-Aware Word Contextualization}

\author{
  Wonseok Hwang \quad
  Jinyeong Yim \quad 
  Seunghyun Park \quad
  Minjoon Seo \quad 
  \\
  Clova AI, NAVER Corp. \\
  \texttt{\{wonseok.hwang, jinyeong.yim, seung.park, minjoon.seo\}@navercorp.com} \\
}

\begin{document}
\maketitle

\begin{abstract}
We present \ours, the first Natural-language-to-SQL (NL2SQL) model to achieve human performance in WikiSQL dataset.
We revisit and discuss diverse popular methods in NL2SQL literature, take a full advantage of BERT~\citep{devlinBERT2018} through an effective table contextualization method, and coherently combine them, outperforming the previous state of the art by 8.2\% and 2.5\% in logical form and execution accuracy, respectively. 
We particularly note that BERT with a seq2seq decoder leads to a poor performance in the task, indicating the importance of a careful design when using such large pretrained models. 
We also provide a comprehensive analysis on the dataset and our model, which can be helpful for designing future NL2SQL datsets and models.  We especially show that our model's performance is near the upper bound in WikiSQL, where we observe that a large portion of the evaluation errors are due to wrong annotations, and our model is already exceeding human performance by 1.3\% in execution accuracy.
\end{abstract}
\section{Introduction}

\noindent 
NL2SQL is a popular form of semantic parsing tasks that asks for translating a natural language (NL) utterance to a machine-executable SQL query.
As one of the first large-scale (80k) human-verified semantic parsing datasets, WikiSQL~\citep{zhongSeq2SQL2017} has attracted much attention in the community and enabled a significant progress through task-specific end-to-end neural models~\citep{xuSQLNet2017}.
On the other side of the NLP community, we have also observed a rapid advancement in contextualized word representations~\citep{peters2018elmo,devlinBERT2018},  which have proved to be extremely effective for most language tasks that deal with unstructured text data.
However, it has not been clear yet whether the word contextualization is also similarly effective when structured data such as tables in WikiSQL are involved. 

In this paper, we discuss our approach on WikiSQL that coherently brings previous NL2SQL liteature and large pretrained models  together. Our model, \ours, consists of two layers, encoding layer that obtains table-aware word contextualization and NL2SQL layer that generates the SQL query from the contextualized representations.
We show that \ours\ outperforms the previous best model achieving 83.6\% logical form accuracy and 89.6\% execution accuracy on WikiSQL test set, outperforming the previous best model by 8.2\% and 2.5\%, respectively. It is important to note that, while BERT plays a significant role, merely attaching a seq2seq model on the top of BERT leads to a poor performance, indicating the importance of properly and carefully utilizing BERT when dealing with structured data.

We furthermore argue that these scores are near the upper bound in WikiSQL, where we observe that most of the evaluation errors are caused by either wrong annotations by humans or the lack of given information. In fact, according to our crowdsourced statistics on an approximately 10\% sampled set of WikiSQL dataset, our model's score exceeds human performance at least by 1.3\% in execution accuracy.

In short, our key contributions are:
\begin{itemize}
    \item We propose a carefully designed architecture that brings the best of previous NL2SQL approaches and large pretrained language models together. Our model clearly outperforms the previous best model and the human performance in WikiSQL.
    \item We provide a diverse and detailed analysis on the dataset and our model. These examinations will further help future research on NL2SQL data creation and model development.
\end{itemize}

The rest of the paper is organized as follows. We first describe our model in Section~\ref{sec:model}. Then we report the quantitative results of our model in comparison to previous baselines in Section~\ref{sec:exp}. Lastly, we discuss qualitative analysis on both the dataset and our model in Section~\ref{subsec:error_analysis}.
\footnote{The source code and human evaluation data is available from \oururl.}

\section{Related Work}

WikiSQL is a large semantic parsing dataset consisting of 80,654 natural language utterances and corresponding SQL annotations on 24,241 tables extracted from Wikipedia~\citep{zhongSeq2SQL2017}.
The task is to build the model that generates SQL query for given natural language question on single table and table headers without using contents of the table. Some examples, using the table from WikiSQL, are shown in Figure~\ref{tab:wikisql_task}.

\begin{figure}[tb]
\scriptsize
\centering
{\bf\sffamily Table:}
  \begin{center}
  \begin{tabular}{lccr}
  \toprule
  Player & \textcolor{blue}{Country} &  \textcolor{red}{Points} & Winnings (\$) \\
  \midrule
  Steve Stricker & United States & 9000 & 1260000 \\
  K.J. Choi & \textcolor{cyan}{South Korea}   & {\bf 5400} & 756000 \\
  Rory Sabbatini & South Africa  & 3400 & 4760000 \\
  Mark Calcavecchia & United States  & 2067 & 289333 \\
  Ernie Els & South Africa & 2067 & 289333 \\
  \bottomrule
  \end{tabular}
  \end{center}
 
{\bf\sffamily Question: } {\sffamily What is the points of South Korea player?} \\
{\bf\sffamily SQL: }{\sffamily {\ttfamily SELECT} \textcolor{red}{Points} {\ttfamily WHERE} \textcolor{blue}{Country} = \textcolor{cyan}{South Korea}} \\
{\bf\sffamily Answer: }{\sffamily 5400}
\caption{Example of WikiSQL semantic parsing task. For given questions and table headers, the model generates corresponding SQL query and retrieves the answer from the table.}
\label{tab:wikisql_task}
\end{figure}

\setlength{\textfloatsep}{5.0pt}

The large size of the dataset has enabled adopting deep neural techniques for the task and drew much attention in the community recently.
Although early studies on neural semantic parsers have started without syntax specific constraints on output space~\citep{dong2016Neural_semantic_parsing,jia2016neuralSP_dataAug,iyer2017UserInLoop}, many state-of-the-art results on WikiSQL have achieved by constraining the output space with the SQL syntax.
The initial model proposed by~\citep{zhongSeq2SQL2017} independently generates the two components of the target SQL query, \texttt{select}-clause and \texttt{where}-clause, which outperforms the vanilla sequence-to-sequence baseline model proposed by the same authors. 
%
SQLNet~\citep{xuSQLNet2017} further simplifies the generation task by introducing a sequence-to-set model in which only \texttt{where} condition value is generated by the sequence-to-sequence model. 
TypeSQL~\citep{yu2018TypeSQL} also employs a sequence-to-set structure but with an additional ``type" information of natural language tokens. 

Coarse2Fine~\citep{dongC2F2018} first generates rough intermediate output, and then refines the results by decoding full \texttt{where}-clauses. 
Also, the table-aware contextual representation of the question is generated with bi-LSTM with attention mechanism which increases logical form accuracy by 3.1\%. Our approach differs in that many layers of self-attentions \citep{vaswani2017transformer,devlinBERT2018} are employed with a single concatenated input of question and table headers for stronger contextualization of the question.

Pointer-SQL~\citep{wang2017pointingOut} proposes a sequence-to-sequence model that uses an attention-based copying mechanism and a value-based loss function.
Annotated Seq2seq~\citep{wang2018srllike} utilizes a sequence-to-sequence model after automatic annotation of input natural language.
MQAN~\citep{mcCann2018decaNLP} suggests a multitask question answering network that jointly learns multiple natural language processing tasks using various attention mechanisms. 
Execution guided decoding is suggested in~\citep{wang2018executionguided}, in which non-executable (partial) SQL queries candidates are removed from output candidates during decoding step.
IncSQL~\citep{shi2018IncSQL} proposes a sequence-to-action parsing approach that uses incremental slot filling mechanism with feasible actions from a pre-defined inventory.

\section{Model}  \label{sec:model}
Our model, \ours, consists of two layers: encoding layer that obtains table- and context-aware question word representations (Section~\ref{subsec:encoding}), and NL2SQL layer that generates the SQL query from the encoded representations (Section~\ref{subsec:nl2sql}).

\subsection{Table-aware Encoding Layer}\label{subsec:encoding}
\begin{figure}[tpb] 
	\centering
	\includegraphics[width=0.5\columnwidth]{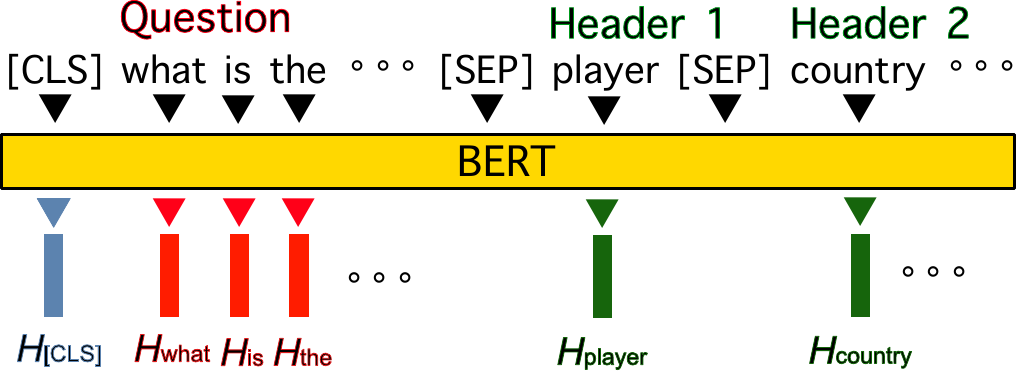}
	\caption{
	(A) The scheme of input encoding process by table-aware BERT. Final output vectors are represented by colored bars: light blue for \texttt{[CLS]} output, red for question words, and green for tokens from table headers.
	}
	\label{fig_sqlova}
\end{figure}

We extend BERT~\citep{devlinBERT2018} for encoding the natural language query together with the headers of the entire table. We use \texttt{[SEP]}, a special token in BERT, to separate between the query and the headers. That is, each query input $T_{n,1} \dots T_{n,L}$ ($L$ is the number of query words) is encoded as \\\\
\texttt{[CLS]}, $T_{n,1}$, $\cdots$ $T_{n,L}$, \texttt{[SEP]}, $T_{h_1, 1}$, $T_{h_1, 2}$, $\cdots$,  \texttt{[SEP]}, $\cdots$, \texttt{[SEP]}, $T_{h_{N_h},1}$, $\cdots$, $T_{h_{N_h}, M_{N_h}}$,\texttt{[SEP]} \\ 
%
%



%
%
\noindent where $T_{h_j, k}$ is the $k$-th token of the $j$-th table header,
$M_j$ is the total number of tokens of the $j$-th table headers,
and $N_h$ is the total number of table headers.
Another input to BERT is the segment id, which is either 0 or 1.
We use 0 for the question tokens and 1 for the header tokens.
Other configurations largely follow~\citep{devlinBERT2018}.
The output from the final two layers of BERT are concatenated and used in \modelBtoM~(Section~\ref{subsec:nl2sql}).
\subsection{NL2SQL Layer}\label{subsec:nl2sql}

\begin{figure*}[t!] 
	\includegraphics[width=1.0\textwidth]{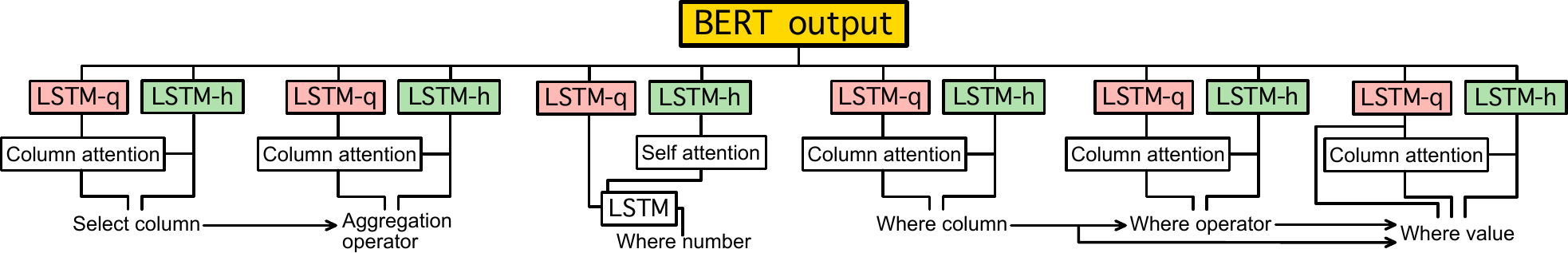}
	\caption{
	The illustration of \modelBtoM\ (Section~\ref{subsec:nl2sql}). The outputs from table-aware encoding layer are encoded again with \texttt{LSTM-q} (question encoder) and \texttt{LSTM-h} (header encoder).
	}
	\label{fig_models_all}
\end{figure*}

In this section, we describe the details of \modelBtoM\ (Figure~\ref{fig_models_all}) on top of the table-aware encoding layer.
%
%

In a typical sequence generation model, the output is not explicitly constrained by any syntax, which is highly suboptimal for formal language generation.
Hence, following~\citep{xuSQLNet2017}, \modelBtoM\ uses syntax-guided sketch, where the generation model consists of six modules, namely \texttt{select-column, select-aggregation, where-number, where-column, where-operator,} and \texttt{where-value} (Figure~\ref{fig_models_all}).
Also, following \citep{xuSQLNet2017}, column-attention is frequently used to contextualize question.

In all sub-modules, the output of table-aware encoding layer (Section~\ref{subsec:encoding}) is further contextualized by two layers of bidirectional LSTM layers with 100 dimension. We denote the LSTM output of the $n$-th token of the question with $E_n$. Header tokens are encoded separately and the output of final token of each header from LSTM layer is used. $D_c$ is used to denotes the encoding of header $c$. The role of each sub-module is described below.

\paragraph{\texttt{select-column}} finds \texttt{select} column from given natural language utterance by contextualizing question through column-attention mechanism~\citep{xuSQLNet2017}.
\bea{
    s(n|c) &= D_c^T \mcW E_n
    \\ p(n|c) &= \softmax( s(n|c) )
    \\ C_c &= \sum_n p(n|c) E_n
    \\ s_{sc}(c) &= \mcW \tanh ( [\mcW D_c; \mcW C_c])
    \\ p_{sc}(c) &= \softmax(s_{sc}(c))
}
Here, $\mcW$ stands for affine transformation, $C_n$ is context vector of question for given column,  $[\cdot;\cdot]$ denotes the concatenation of two vectors, and $p_{sc}(c)$ indicates the probability of generating column $c$. To make the equation uncluttered, same $\mcW$ is used to denote any affine transformation in our paper although all of them denote different transformations.

\paragraph{\texttt{select-aggregation}} finds aggregation operator \texttt{agg} for given column $c$ among six possible choices (\texttt{NONE}, \texttt{MAX}, \texttt{MIN}, \texttt{COUNT}, \texttt{SUM}, and \texttt{AVG}). Its probability is obtained by
\bea{
    p_{sa}(\texttt{agg} | c ) = \softmax( \mcW \tanh \mcW C_c)
}
where $C_c$ is the context vector of the question obtained by the same way in \texttt{select-column}.

\paragraph{\texttt{where-number}} finds the number of \texttt{where} condition by contextualizing column ($C$) via self-attention and contextualizing question ($C_Q$) conditioned on $C$.
\bea{
    p(c) &= \softmax( \mcW D_c)
    \\ C &= \sum_c p(c) D_c
    \\ h &= \mcW C
    \\ c &= \mcW C
    \\ p'(n) &= \softmax (\mcW ~ \text{bi-LSTM}( E_n, h, c))
    \\ C_{Q} &= \sum_n p'(n) E_n
    \\ s_{wn} &= \mcW \tanh \mcW C_Q
}
Here, $h$ and $c$ are initial ``hidden" and ``cell" inputs to LSTM encoder.
The probability of observing $k$ number of \texttt{where} condition is found from $k$-th element of vector $\softmax( s_{wn} )$. 
This submodule is same with that of SQLNet~\citep{xuSQLNet2017} and shown here for comprehensive reading.

\paragraph{\texttt{where-column}} obtains the probability of observing column $c$ ($p_{wc}(c)$) through column-attention,
\bea{
    s_{wc}(c) &= \mcW \tanh( [\mcW D_c ; \mcW C_c])
    \\ p_{wc}(c) &= \text{sigmoid}(s_{wc}(c))
}
where $C_c$ is the context vector of the question obtained by the same way as in \texttt{select-column}.
The probability of generating each column is obtained separately from \texttt{sigmoid} function and top $k$ columns are selected. $k$ is found from \texttt{where-number} sub-module.

\paragraph{\texttt{where-operator}} finds \texttt{where} operator \texttt{op} ($\in \{=, >, <\}$) for given column $c$ through column-attention.
\bea{
s_{wo}(\texttt{op}|c) &= \mcW \tanh \mcW ( [\mcW D_c; \mcW C_c] )
\\ p_{wo}(\texttt{op}|c) &= \softmax s_{wo}(\texttt{op} | c)
}
where $C_c$ is the context vector of the question obtained by the same way in \texttt{select-column}.

\paragraph{\texttt{where-value}} finds \texttt{where} condition by locating start- and end-tokens from question for given column $c$ and operator \texttt{op}.
\bea{
\texttt{vec} &= [E_n; \mcW C_c; \mcW D_c; \mcW V_\texttt{op}]
\\s_{wv}(n|c, \texttt{op}) &= \mcW \tanh \mcW ~ \texttt{vec}
}
Here, $V\texttt{op}$ stands for one-hot vector of \texttt{op} ($\in \{=, >, <\}$).
The probability of $n$-th token of question being start-index for given $c$-th column and \texttt{op} is obtained by feeding 1st element of $s_{wv}$ vector to softmax function whereas that of end-index is obtained by using 2nd element of $s_{wv}$ by the same way.

To sum up, our \modelBtoM\ is motivated by SQLNet \citep{xuSQLNet2017} but have following key differences.
Unlike SQLNet, \modelBtoM\ does not share parameters. Also, instead of using pointer network for inferring the \texttt{where} condition values, we train for inferring the start and the end positions of the utterance, following~\citep{dongC2F2018}. Furthermore, the inference of the start and the end tokens in \texttt{where-value} module depends on both selected \texttt{where-column} and \texttt{where-operators} while the inference relies on \texttt{where-columns} only in~\citep{xuSQLNet2017}. Lastly, when combining two vectors corresponding to the question and the headers, concatenation instead of addition is used.

\paragraph{Execution-Guided Decoding (EG)}
During the decoding (SQL query generation) stage, non-executable (partial) SQL queries can be excluded from the output candidates for more accurate results, following the strategy suggested by~\citep{wang2018executionguided,yin2018TRANX}.
In \texttt{select} clause, (\texttt{select} column, aggregation operator) pairs are excluded when the string-type columns are paired with numerical aggregation operators such as \texttt{MAX}, \texttt{MIN}, \texttt{SUM}, or \texttt{AVG}.
The pair with highest joint probability is selected from remaining pairs.
In \texttt{where} clause decoding, the executability of each (\texttt{where} column, operator, value) pair is tested by checking the answer returned by the partial SQL query \texttt{select agg(col$_{s}$) where col$_{w}$ op val}. Here, \texttt{col$_s$} is the predicted \texttt{select} column, \texttt{agg} is the predicted aggregation operator, \texttt{col$_w$} is one of the \texttt{where} column candidates, \texttt{op} is \texttt{where} operator, and \texttt{val} stands for the \texttt{where} condition value. 
 The queries with empty returns are also excluded from the candidates. 
 The final output of \texttt{where} clause is determined by selecting the output maximizing the joint probability estimated from the output of \texttt{where-number}, \texttt{where-column}, \texttt{where-operator}, and \texttt{where-value} modules.


\section{Experiments}\label{sec:exp}

During training, BERT-based table-aware encoding layer (\texttt{BERT-Large-Uncased}\footnote{https://github.com/google-research/bert}) are loaded and fine-tuned with ADAM optimizer with the learning rate of $10^{-5}$, whereas \modelBtoM\ is trained with the learning rate of $10^{-3}$. In both cases, the decay rates of ADAM optimizer are set to $\beta_1 = 0.9, \beta_2 = 0.999$.
Batch size is set to 32.
To find word vectors, natural language utterance is first tokenized by using Standford CoreNLP~\citep{manningCoreNLP2014}.
Each token is further tokenized (into sub-word level) by WordPiece tokenizer~\citep{devlinBERT2018,wu2016word_piece}.
The headers of the tables and SQL vocabulary are tokenized by WordPiece tokenizer directly.
The PyTorch version of BERT code\footnote{https://github.com/huggingface/pytorch-pretrained-BERT} is used for word embedding and some part of the code in~\modelBtoM~is influenced by the original SQLNet source code\footnote{https://github.com/xiaojunxu/SQLNet}.
All experiments were performed on WikiSQL ver. 1.1 \footnote{https://github.com/salesforce/WikiSQL}.

\subsection{Accuracy Measurement}
The logical form (LF) and the execution accuracy (X) on dev set (consisting of 8,421 queries) and test set (consisting of 15,878 queries) of WikiSQL of several models are shown in Table~\ref{tbl_sota}. The execution accuracy is measured by evaluating the answer returned by `executing' the query on the SQL database. The order of \texttt{where} conditions is ignored in measuring logical form accuracy in our models.
The top rows in Table~\ref{tbl_sota} show models without execution guidance (EG), and the bottom rows show models augmented with EG.
\ours\ outperforms previous baselines by a large margin,
achieving [+5.3\% LF] and [+2.5\% X] for non-EG and 
achieving [+8.2\% LF] and [+2.5\% X] for EG. 

\begin{table*}[t!] 
\caption{Comparison of various models. Logical from accuracy (LF) and execution accuracy (X) on dev and test set of WikiSQL. ``EG" stands for ``execution-guided".}
\label{tbl_sota}
\footnotesize
\begin{threeparttable}
\begin{tabular*}{1.0\textwidth}{l@{\extracolsep{\fill}}cccr}
\toprule
Model & Dev LF (\%) & Dev X (\%) & Test LF (\%) & Test X (\%) \\
\midrule
Baseline~\citep{zhongSeq2SQL2017}  & 23.3 & 37.0 &  23.4& 35.9  \\
Seq2SQL~\citep{zhongSeq2SQL2017} &49.5 & 60.8 & 48.3 & 59.4  \\ 
SQLNet~\citep{xuSQLNet2017}  & 63.2 & 69.8 & 61.3 & 68.0 \\ 
PT-MAML~\citep{pshuang2018PTMAML} & 63.1 & 68.3 & 62.8 & 68.0 \\ 
TypeSQL~\citep{yu2018TypeSQL} & 68.0 & 74.5 & 66.7 & 73.5 \\ 
Coarse2Fine~\citep{dongC2F2018} & 72.5 & 79.0 & 71.7  & 78.5 \\ 
MQAN~\citep{mcCann2018decaNLP} & 76.1 & 82.0 & 75.4 & 81.4 \\ 
Annotated Seq2seq~\citep{wang2018srllike} \footnotemark[1] & 72.1 & 82.1 & 72.1 & 82.2 \\ 
IncSQL~\citep{shi2018IncSQL} \footnotemark[1] &  49.9 & 84.0 & 49.9 & 83.7 \\ 
\midrule
\modelBaseline\ (ours) & 57.3 & - & 56.4 & -\\
\modelBaselineT\ (ours) & 70.5 & - & - & -\\
\ours\ (ours) &\textbf{81.6} (\textbf{+5.5}) & {\bf 87.2} ({\bf +3.2}) &\textbf{ 80.7} (\textbf{+5.3}) & {\bf 86.2} ({\bf +2.5}) \\ 
\midrule
PointSQL+EG~\citep{wang2018executionguided} \footnotemark[1]$^{,}$\footnotemark[2]& 67.5 & 78.4 & 67.9 & 78.3 \\ 
Coarse2Fine+EG~\citep{wang2018executionguided} \footnotemark[1]$^{,}$\footnotemark[2] & 76.0 & 84.0 & 75.4 & 83.8 \\ 
IncSQL+EG~\citep{shi2018IncSQL} \footnotemark[1]$^{,}$\footnotemark[2]& 51.3 & 87.2 & 51.1 & 87.1 \\ 
\midrule
\ours+EG (ours) \footnotemark[2] &\textbf{84.2} (\textbf{+8.2}) &\textbf{90.2} (\textbf{+3.0})& \textbf{83.6} (\textbf{+8.2})& \textbf{89.6 (+2.5)}\\ 
\midrule
Human performance \footnotemark[3] & - & - & - & 88.3\\ 
\bottomrule
\end{tabular*}
\begin{tablenotes}
\footnotesize
\item[1]{Source code is not opened.}
\item[2]{Execution guided decoding is employed.}
\item[3]{Measured over 1,551 randomly chosen samples from WikiSQL test set (Section~\ref{subsec:error_analysis}).}
\end{tablenotes}
\end{threeparttable}
\end{table*}
\setlength{\textfloatsep}{0.3cm}

To understand the performance of \ours\ in detail, the logical form accuracy of each sub-module was obtained and shown in Table~\ref{tbl_acc_submodule}.
All sub-modules show~$\small\GtrSim$~95\% in accuracy except \texttt{select-aggregation} module whose low accuracy partially results from the error in the ground-truth of WikiSQL (Section-\ref{subsec:error_analysis}).\footnote{In addition to \ours\ we also provide two BERT-based models which also outperforms previous baselines by large margin, in  Appendix \ref{subsec:extra_models}.} 
\begin{table*}[th!]
\centering
\caption{The logical from accuracy of each sub-module over WikiSQL dev set.
\texttt{s-col, s-agg, w-num, w-col, w-op} and \texttt{w-val} stand for \texttt{select-column, select-aggregation, where-number, where-column, where-operstor}, and \texttt{where-value} respectively.
}
\label{tbl_acc_submodule}
	\footnotesize
\begin{tabular*}{1.0\textwidth}{l @{\extracolsep{\fill}} cccccr}
\toprule
Model & \texttt{s-col} & \texttt{s-agg}  & \texttt{w-num} & \texttt{w-col} & \texttt{w-op} & \texttt{w-val} \\
\midrule
\ours, Dev  & 97.3 & 90.5 & 98.7 & 94.7& 97.5& 95.9\\
\ours, Test & 96.8 & 90.6 & 98.5 & 94.3& 97.3& 95.4\\ 
\bottomrule
\end{tabular*}
\end{table*}

Choosing not to answer for low-confidence predictions is another important measure of performance. 
We use the output probability of a generated SQL query from \ours\ as the confidence score and predict that the question is \emph{unanswerable} when the score is low.
%
The result shows that \ours\ effectively assigns a low probability to wrong predictions, yielding a high precision of 95\%+ with a recall rate of 80\%.
The precision-recall curve and its area under curve are shown in Figure~\ref{fig:auc}. 

\subsection{Ablation Study}
To understand the importance of each part of \ours, we evaluate ablations in Table~\ref{tbl_abl}.
The results show that word contextualization (without fine-tuning) contributes to overall logical form accuracy by 4.1\% (dev) and 3.9\% (test) (compare third and fifth rows of the table) which is similar to the observation by~\citep{dongC2F2018} where the 3.1\% increases observed with table-aware LSTM encoder. Consistently, replacing BERT by ELMo \citep{peters2018elmo} shows similar results (fourth row of the table).
But unlike GloVe, where fine-tuning increases only a few percents in accuracy \citep{xuSQLNet2017}, fine-tuning of BERT increases the accuracies by 11.7\% (dev) and 12.2\% (test) (compare first and third rows in the table) which may be attributed to the use of many layers of self-attention \citep{vaswani2017transformer}. Use of \texttt{BERT-Base} decreases the accuracy by 1.3\% on both dev and test set compared to \texttt{BERT-Large} cases.
We also developed \modelBaseline\ where the encoder part of vanilla sequence-to-sequence model with attention \citep{jia2016neuralSP_dataAug} is replaced by BERT. The model achieves 57.3\% and 56.4\% logical form accuracies in dev and test sets respectively (Table \ref{tbl_sota}) highlighting the importance of using proper decoding layers. To further validate the conclusion, we replaced LSTM decoder in \modelBaseline\ into Transformer (\modelBaselineT), the model achieved 70.5 logical form accuracy in dev set again achieving 11.1\% lower score compared to \ours.
The detailed description of the model is presented in Appendix \ref{subsec:bs2s}.

\begin{table*}[ht!] 
\centering
\caption{ The results of ablation study. Logical from accuracy (LF) and execution accuracy (X) on dev and test sets of WikiSQL are shown.}
\footnotesize
\begin{tabular*}{1.0\textwidth}{l @{\extracolsep{\fill}} cccr}
\toprule
 Model & Dev LF (\%) & Dev X (\%) &Test LF (\%)&Test X (\%)\\
\midrule
     \ours\ &81.6 &  87.2  & 80.7  & 86.2  \\ 
     ~~~~~(-) BERT-Large (+) BERT-Base  &80.3 & 85.8 & 79.4 & 85.2 \\
     ~~~~~(-) Fine-tuning &  69.9 & 77.0 & 68.5 & 75.6 \\ 
     ~~~~~(-) BERT-Large (+) ELMo (fine-tuned) &  71.3 & 77.7 & 69.6 & 76.0 \\ 
     ~~~~~(-) BERT-Large (+) GloVe &  65.8 & 72.9 & 64.6  & 71.7 \\
\bottomrule
\end{tabular*}
\label{tbl_abl}
\end{table*}

\section{Analysis} \label{subsec:error_analysis}
\newcommand{\STAB}[1]{\begin{tabular}{@{}c@{}}#1\end{tabular}}

\subsection{Error Analysis}

There are 1,533 mismatches in logical form between the ground-truth (GT) and the predictions from \ours\ in WikiSQL dev set. 
Among the mismatches,
100 samples were randomly selected, analyzed, and classified into two categories: (1) 26 ``unanswerable'' cases of which it is not possible to generate correct SQL query for given information (question and table schema), and (2) 74 ``answerable'' cases.

Unanswerable cases were further categorized into the following four types.
\begin{itemize}
\item Type I: the headers of tables do not contain the necessary information.
For example, a question ``{\small\ttfamily What was the score between Marseille and Manchester United on the second leg of the Champions League Round of 16?}''  and its corresponding table headers {\small$\{$`{\ttfamily Team}', `{\ttfamily Contest and round}', `{\ttfamily Opponent}', `{\ttfamily 1st leg score}', `{\ttfamily 2nd leg score}', `{\ttfamily Aggregate score}'$\}$} in QID-1986 (Table~\ref{tab:100example}) do not contain information about which header should be selected for condition values `{\small\ttfamily Manchester United}' and `{\small\ttfamily Marseille}'. 
\item Type II: There exist multiple valid SQL queries per question (QID-783, 2175, 4229 in Table \ref{tab:100example}). For example, the GT SQL query of QID-783 has \texttt{count} aggregation operator and any header can be used for \texttt{select} column.
%
\item Type III: the generation of nested SQL query is required. For example, correct SQL query for QID-332 (Table \ref{tab:100example}) is ``{\small\ttfamily SELECT count(incumbent) WHERE District=(SELECT District WHERE Incumbent=Alvin Bush)}''.
\item Type IV: questions are ambiguous. For example, the answer to the question ``{\small\ttfamily What is the number of the player who went to Southern University?}'' in QID-156 (Table \ref{tab:100example}) can vary depending on the interpretation of ``{\small\ttfamily the number of the player}''.
\end{itemize}
The categorization of 26 samples is summarized in Table~\ref{tab:error_type} in Appendix~\ref{app:unanswerable}..

\begin{figure}[th]
	\centering
	\includegraphics[width=1\columnwidth]{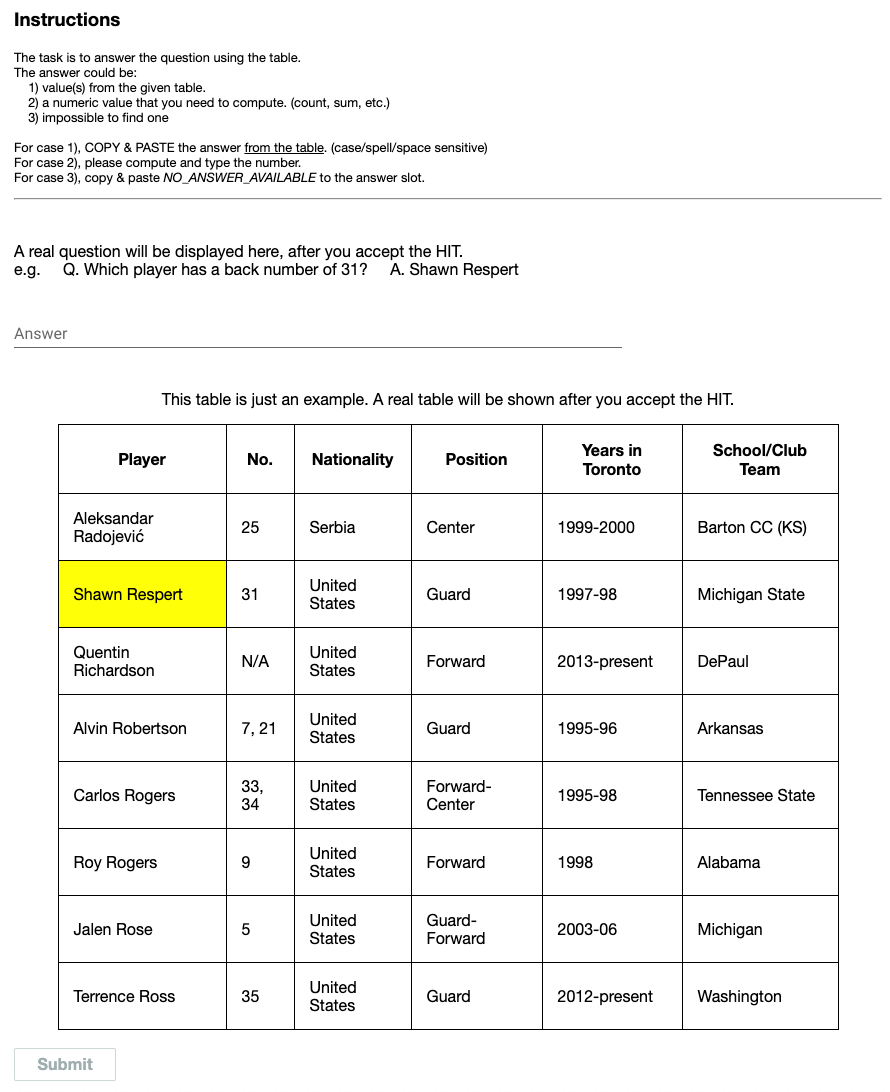}
	\caption{The instruction and example given to crowdworkers during human performance evaluation on the WikiSQL dataset}
	\label{fig:amt}
\end{figure}

Further analysis over the remaining 74 answerable examples reveals that there are 49 GT errors in logical forms. 
45 out of 49 examples contain GT errors in aggregation operators (e.g. QID-7062), two have GT errors in \texttt{select} columns (e.g. QID-841, 5611), and remaining two contain GT errors in \texttt{where} clause (e.g. QID-2925, 7725).
Interestingly, among 49 examples, 41 logical forms are correctly predicted by \ours, indicating that the actual performances of the models in Table~\ref{tbl_sota} are underestimated. This also may imply that most of examples in WikiSQL have correct GT for training. The results are summarized in Table~\ref{tab:dataset_err}, and all 100 examples are presented in Table~\ref{tab:100example} in Appendix~\ref{app:100examples}. 

As the questions in WikiSQL are created by paraphrasing queries generated automatically from the templates without considering the table contents, the meanings of the questions could change, especially when the quantitative answer is required, possibly leading to GT errors. For example, QID-3370 in Table~\ref{tab:100example} is related to an ``{\small\ttfamily year}'' and the GT SQL query includes unnecessary {\ttfamily COUNT} aggregation operators.

Overall, the error analysis above may imply that near-$90\%$ accuracy of \ours\ could be near the upper bound in WikiSQL task the ``answerable'' and non-erroneous questions when the contents of tables are not available. 
%

%
%





\subsection{Measuring Human Performance} \label{subsec:human_performance}

The human performance on WikiSQL dataset has not been measured so far despite its popularity.
Here, we provide the approximate human performance by collecting answers from 246 different crowdworkers through Amazon Mechanical Turk over 1,551 randomly sampled examples from the WikiSQL test set (which has 15,878 examples in total). The crowdworkers were selected with following three constraints: (1) 95\% or higher task acceptance rate; (2) 1000 or higher HITs; (3) residents of the United States. 

During the evaluation, crowdworkers were asked either to find value(s) or to compute a value using the given questions and corresponding tables following the instruction provided (Figure \ref{fig:amt}).
Note that the task requires general capability of understandings English text and finding values from a table without a need for the generation of SQL queries.
This effectively mimics the measurement of execution accuracy in WikiSQL. We find that the accuracy of crowdworkers on the randomly sampled test data is 88.3\%, as shown in Table~\ref{tbl_sota} while the execution accuracy of \ours\ over 1,551 samples are 86.8\% (w/o EG) and 91.0\% (w/ EG). \footnote{When measuring human performance, errors in ground truth are manually corrected by experts (us).} 

We manually checked and analyzed all answers from the crowd. Errors made by crowdworkers are similar to that of the model such as a mismatch of \texttt{select} columns or \texttt{where} columns. One notable mistake by only humans (that our model does not make) is confusion on the ambiguity of natural language. For example, when a question is asking a column value with more than two conditions, crowdworkers show the tendency to consider a single condition only because multiple conditions were written with ``and" which is often considered as the meaning of ``or" in real life.

\section{Conclusion}

In this paper, we propose the first NL2SQL model to achieve a super-human accuracy in WikiSQL. We demonstrate the effectiveness of a careful architecture design that brings and combines previous approaches in NL2SQL and table-aware word contextualization with large pretrained language model (BERT) together. We propose a BERT-based table-aware encoder and a task-specific module on the top of the encoder, outperforming the previous best model by 8.2\% and 2.5\% in logical form and execution accuracy, respectively. 
We hope our detailed explanation and analysis of the model and the dataset provide an insight on how future research on NL2SQL models and datasets can be effectively approached.


\subsubsection*{Acknowledgments}
We thank Clova AI members for their great support, especially Jung-Woo Ha for proof-reading the manuscript, Sungdong Kim and Dongjun Lee for providing help on using BERT, Guwan Kim for insightful comments. We also thank the Hugging Face Team for sharing the PyTorch implementation of BERT.

\medskip
\small
\bibliography{./semantic_parsing.bib}
\bibliographystyle{./acl_natbib.bst}

\appendix

\onecolumn
\section{Appendix}
\setcounter{figure}{0}  
\setcounter{equation}{0}
\renewcommand{\thefigure}{A\arabic{figure}} 
\renewcommand{\theequation}{A\arabic{equation}}

\subsection{Additional models} \label{subsec:extra_models}
\subsubsection{\modelBtoScal}
\begin{figure*}[t!] 
	\includegraphics[width=1.0\textwidth]{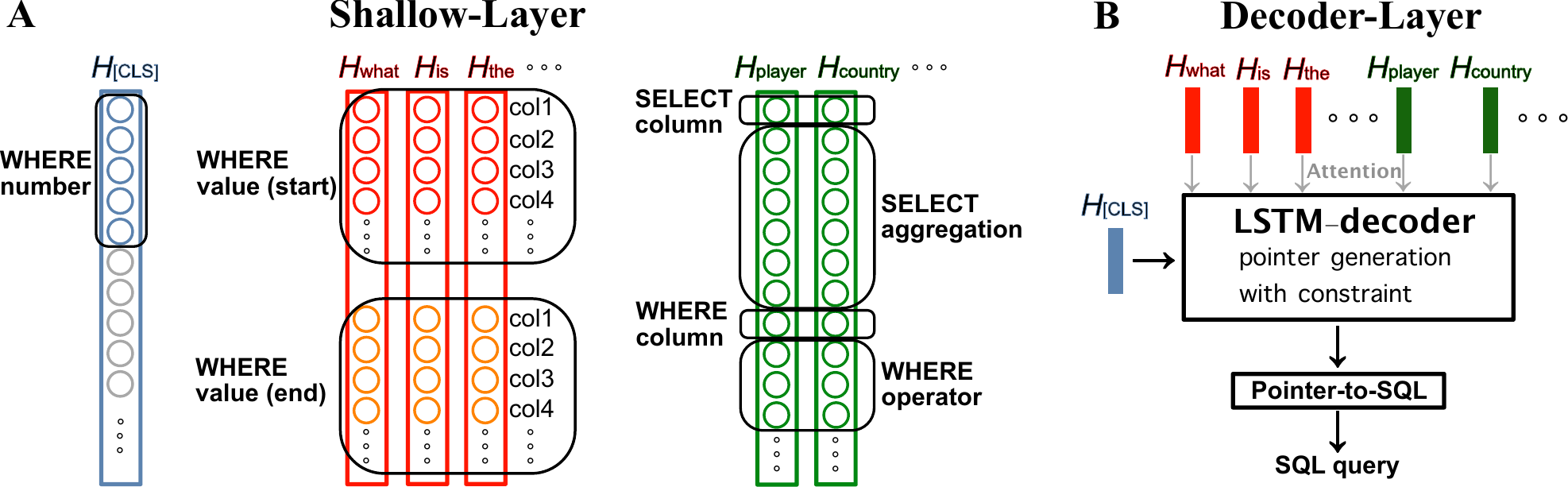}
	\caption{
	(A) The model scheme of \modelBtoScal. Each circle represents the single element of the output vector from table-aware BERT-encoder. The circles are grouped by black squares according to their roles in SQL query generation.
	(B) The scheme \modelBtoSeq. \texttt{LSTM-docoder} of pointer network \citep{vinyals2015NIPS_pointerNetwork} generates the sequence of pointers to augmented inputs which include SQL vocabulary, \texttt{start}, \texttt{end}, question words, and header tokens. Generated pointer squences are interpreted by \texttt{Pointer-to-SQL} module which generates final SQL queries.
	}
	\label{fig_models_all2}
\end{figure*}

Here, we present another task specific layer \modelBtoScal\ having lower model complexity compared to \modelBtoM. \modelBtoScal\ does not contain trainable parameters but controls the flow of information during fine-tuning of BERT via loss function.
Like \modelBtoM, \modelBtoScal\ uses syntax-guided sketch, where the generation model consists of six modules, namely \texttt{select-column, select-aggregation, where-number, where-column, where-operator,} and \texttt{where-value} (Figure~\ref{fig_models_all2}A). 

%

\texttt{select-column} module finds the column in \texttt{select} clause from given natural language utterance by modeling the probability of choosing $i$-th header ($p_{sc}(\texttt{col}_i)$) as
\bea{
p_{sc}(\texttt{col}_i) &= \softmax (H_{h,i})_0
}
where $H_{h,i}$ is the contextualized output vector of first token of $i$-th header by table-aware BERT encoder, and $(H_{h,i})_0$ indicates zeroth element of the vector $H_{h,I}$. In general, $(V)_\mu$ denotes $\mu$-th element of vector $V$ in this paper. Also, the conditional probability for given question and table-schema $p(\cdot|\text{Q}, \text{table-schema})$ is simply denoted as $p(\cdot)$ to make equation uncluttered.

\texttt{select-aggregation} module finds the aggregation operator for the given \texttt{select} column. The probability of generating aggregation operator \texttt{agg} for given \texttt{select} column $\texttt{col}_i$ is described by 
\bea{
p_{sa}(\texttt{agg}_\mu|\texttt{col}_i) &= \softmax \pave{(H_{h,i})_\mu}
}
where $\texttt{agg}_1$, $\texttt{agg}_2$, $\texttt{agg}_3$, $\texttt{agg}_4$, $\texttt{agg}_5$, and $\texttt{agg}_6$ are \texttt{none}, \texttt{max}, \texttt{min}, \texttt{count}, \texttt{sum}, and \texttt{avg} respectively.

\texttt{where-number} module predicts the number of \texttt{where} conditions by modeling the probability of generating $\mu$-number of conditions as
\bea{
    p_{wn}(\mu) &= \softmax \pave{(\mcW H_{\texttt{[CLS]}})_\mu}
}
where $H_{\texttt{[CLS]}}$ is the output vector of \texttt{[CLS]} token from table-aware BERT encoder, and $\mcW$ stands for affine transformation. Throughout the paper, any affine transformation shall be denoted by $\mcW$ for the clarity.

\texttt{where-column} module calculates the probability of generating each columns in \texttt{where} clause. The probability of generating $\texttt{col}_i$ is given by
\bea{
    p_{wc}({\text{col}_i}) &= \text{sigmoid}((H_{h,i})_7).
}

\texttt{where-operator} module finds most probable operators for given \texttt{where} column among three possible choices ($>, =, <$). The probability of generating operator $\texttt{op}_\mu$ for given \texttt{where} column $\texttt{col}_i$ is modeled as
\bea{
    p_{wo}(\texttt{op}_\mu | \texttt{col}_i) 
    &= \softmax \pave{ (H_{h, i})_\mu }
}
where $\texttt{op}_8$, $\texttt{op}_9$, and $\texttt{op}_{10}$ are \texttt{>}, \texttt{=}, and \texttt{<} respectively.

\texttt{where-value} module finds which tokens of a question correspond to condition values for given \texttt{where} columns by locating start- and end-tokens.
The probability that $k$-th question token is selected as a start token for given \texttt{where} column $\texttt{col}_\mu$ is modeled as
\bea{
p_{wv,st}(k | \texttt{col}_\mu) 
&= \softmax \pave{ (H_{n,k})_\mu }.
}
Similarly the probability of $k$-th question token is selected as an end token is
\bea{
p_{wv,ed}(k | \texttt{col}_\mu) 
&= \softmax \pave{ (H_{n,k})_{\mu+100} }
}
100 is selected to avoid overlap during inference between start- and end-token models. The maximum number of table headers in single table is 44 in WikiSQL task.
\\

\subsubsection{\modelBtoSeq}
\modelBtoSeq\ contains LSTM decoders adopted from pointer network~\citep{vinyals2015NIPS_pointerNetwork,zhongSeq2SQL2017}  (Fig. \ref{fig_models_all}B) with following special features. Instead of generating entire header tokens, we only generate first token of each header and interpret them as entire header tokens during inference stage using \texttt{Point-to-SQL} module (Fig. \ref{fig_models_all}B). Similarly, the model generates only the pointers to start- and end- \texttt{where}-value tokens omitting intermediate points.
Decoding process can be expressed as following equations which use the attention mechanism.
\bea{
    D_t &= \text{LSTM}(P_{t-1}, (h_{t-1}, c_{t-1}))
    \\ h_0 &= (\mcW H_{\texttt{([CLS])}})_{0:d}
    \\ c_0 &= (\mcW H_{\texttt{([CLS])}})_{d:2d}
    \\ s_{t}(i) &=  \mcW ( \mcW H_{i} + \mcW D_t)
    \\ p_{t}(i) &= \softmax ~ s_{t}(i).
}
$P_{t-1}$ stands for the one-hot vector (pointer) at time $t-1$, $h_{t-1}$ and $c_{t-1}$ are hidden- and cell- vectors of LSTM decoder, $d$ is the hidden dimension of BERT, $H_i$ is the BERT output of $i$-th token,  and $p_t(i)$ is the probability observing $i$-th token at time $t$.

\subsubsection{\modelBaseline} \label{subsec:bs2s}
\modelBaseline\ consists of the table-aware BERT encoder and LSTM decoder which is essentially a sequence-to-sequence model (with attention) \citep{jia2016neuralSP_dataAug} except that the LSTM encoder part is replaced by BERT.
The encoding process is same  with \modelBtoM. The decoding process is described by following equations.
\bea{
    E_t &= \text{emb}_{\text{BERT}} (w_t)
    \\ \rho_i(t) &= H_i^T \mcW E_t
    \\ C_t &= \sum_i H_{i} \rho_i(t)
    \\ h_{t+1} &= \text{LSTM}([C_t; E_t], h_t)
    \\ h_0 &= (\mcW H_{\texttt{([CLS])}})_{0:d}
    \\ c_0 &= (\mcW H_{\texttt{([CLS])}})_{d:2d}
    \\ p(w_{t+1}) &= \softmax( \mcW \tanh h_{t+1} [C_t; h_{t+1}] )
}
where $w_t$ stands for word token (among 30,522 token vocabulary used in BERT) predicted at time $t$, $\text{emb}_\text{BERT}$ is a map that transform the token to embedding vector $E_t$ via word embedding module of BERT, $H_i$ is the output vector from BERT encoder of $i$-th input token, and $p(w_{t+1})$ is the probability of generating token $w_{t+1}$ at time $t+1$.

\subsubsection{The performance of \modelBtoScal\ and \modelBtoSeq}

Compared to previous best results, 
\modelBtoScal~shows +5.5\% LF and +3.1\% X,~\modelBtoSeq~shows +4.4\% LF and +1.8\% X
for non-EG case (Table. \ref{tbl_extra}).
For EG case, \modelBtoScal~shows +6.4\% LF and +0.4\% X,~\modelBtoSeq~shows +7.8\% LF and +2.5\% X (Table. \ref{tbl_extra}).

\modelBtoScal~shows [+6.\% LF] and [+3.1\% X] whereas


\begin{table*}[ht!] 
\caption{Logical from accuracy (LF) and execution accuracy (X) on dev and test set of WikiSQL. ``EG" stands for ``execution-guided".}
\label{tbl_extra}
\footnotesize
\begin{threeparttable}
\begin{tabular*}{1.0\textwidth}{l@{\extracolsep{\fill}}cccr}
\toprule
Model & Dev LF (\%) & Dev X (\%) & Test LF (\%) & Test X (\%) \\
\midrule
\modelBtoScal\ (ours)&  \textbf{81.5} (\textbf{+5.4}) & \textbf{87.4} (\textbf{+3.2})& \textbf{80.9} (\textbf{+5.5}) & \textbf{86.8} (\textbf{+3.1})\\ 
\modelBtoSeq\ (ours)&  \textbf{79.7} (\textbf{+3.6}) & \textbf{85.5} (\textbf{+1.1})& \textbf{79.8} (\textbf{+4.4}) & \textbf{85.5} (\textbf{+1.8})\\ 

\midrule
\modelBtoScal-EG (ours)&  \textbf{82.3} (\textbf{+6.3}) & \textbf{88.1} (\textbf{+0.9})& \textbf{81.8} (\textbf{+6.4}) & \textbf{87.5} (\textbf{+0.4})\\ 
\modelBtoSeq-EG (ours)&  \textbf{83.4} (\textbf{+7.4}) & \textbf{89.9} (\textbf{+2.7})& \textbf{83.2} (\textbf{+7.8}) & \textbf{89.6} (\textbf{+2.5})\\ 

\bottomrule
\end{tabular*}

\end{threeparttable}
\end{table*}
\setlength{\textfloatsep}{0.3cm}

\subsection{The Precision-Recall Curve}
\begin{figure}[!h]
	\centering
	\includegraphics[width=0.5\columnwidth]{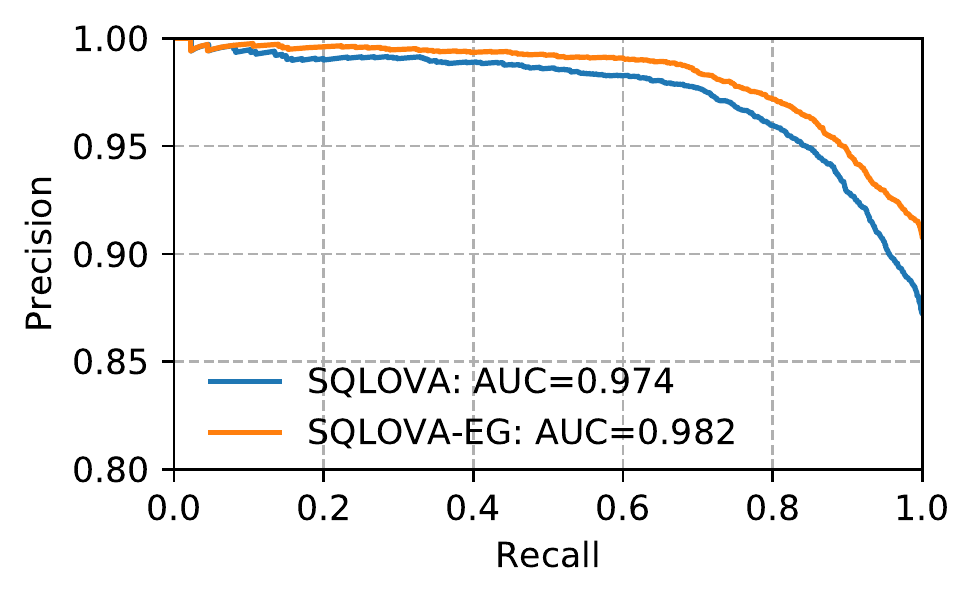}
	\caption{Precision-Recall curve and area under curve (AUC) with SQLova (blue) and SQLova-EG (orange). Precision and recall rates are controlled by varying the threshold value for the confidence score.}
	\label{fig:auc}
\end{figure}

\subsection{The Contingency Table}

\begin{table}[h!]
\centering
\caption{Contingency table of 74 answerable questions. Corresponding 74 ground truth- (GT) and predicted-SQL queries by ~\ours\ are manually classified to correct and incorrect cases.}
\label{tab:dataset_err}
\scriptsize
\begin{tabular}{lcccr}
& & \multicolumn{2}{c}{SQL (GT)} & \\
\cmidrule{3-4}
& & correct & incorrect & total \\
\toprule
\multirow{2}{*}{\parbox{1.0cm}{\centering SQL\\(Ours)}}
 & correct & 0 & \textbf{41} & 41 \\
 \cmidrule{2-5}
& incorrect & 25 & 8 & 33 \\
\midrule & total & 25 & 49 & 74 \\
\bottomrule
\end{tabular}
\end{table}

\subsection{The Types of Unanswerable Examples}\label{app:unanswerable}

%



\begin{table*}[h]
\caption{26 unanswerable examples. ``types'' denotes the type of unanswerable cases that each question belongs to. ``total'' means the number of examples in the type.}
\label{tab:error_type}
\centering
\small
\begin{tabular*}{1.0\textwidth}{lp{12cm}r}
\toprule
types    &   QID  & total \\
\midrule
Type I	&	\hyperlink{app:qid_19}{19}, \hyperlink{app:qid_261}{261}, \hyperlink{app:qid_557}{557}, \hyperlink{app:qid_597}{597}, \hyperlink{app:qid_598}{598}, \hyperlink{app:qid_738}{738}, \hyperlink{app:qid_1089}{1089}, \hyperlink{app:qid_1122}{1122}, \hyperlink{app:qid_1430}{1430}, \hyperlink{app:qid_1986}{1986}, \hyperlink{app:qid_2891}{2891}, \hyperlink{app:qid_3050}{3050}, \hyperlink{app:qid_3602}{3602}, \hyperlink{app:qid_3925}{3925}, \hyperlink{app:qid_5893}{5893}, \hyperlink{app:qid_6028}{6028}, \hyperlink{app:qid_6533}{6533}, \hyperlink{app:qid_7070}{7070}, \hyperlink{app:qid_7912}{7912}, \hyperlink{app:qid_8041}{8041}, \hyperlink{app:qid_8111}{8111}	&	21 \\
\midrule
Type II	&	\hyperlink{app:qid_783}{783}, \hyperlink{app:qid_2175}{2175}, \hyperlink{app:qid_4229}{4229}	&	3 \\
\midrule
Type III	&	\hyperlink{app:qid_332}{332}	&	1 \\
\midrule
Type IV	&	\hyperlink{app:qid_156}{156}	&	1 \\
\bottomrule
\end{tabular*}
\end{table*}

\subsection{100 Examples in the WikiSQL Dataset}\label{app:100examples}

\footnotesize
\begin{longtable}[ht]{lllp{0.73\textwidth}}
\caption{The dataset examples from WikiSQL dev set used in Section~\ref{subsec:error_analysis}. 100 samples were randomly selected from 1,533 mismatches between the ground-truth and the predictions of \ours. QID denotes an index of the question among 8,421 wikiSQL dev set data. There are three types of queries: natural language queries (NL), ground truth SQL queries (SQL (T)), predicted SQL queries (SQL (P)). Other fields indicate ground truth answer (ANS (T)), predicted answer (ANS (P)), and a type of error (ERROR), respectively. Note that the types of unanswerable cases that the question belongs to are shown in the parentheses after ``Question'' in the ``Error'' field.}\label{tab:100example} \\
\toprule
No.	&	QID	&	Type	&	Description \\ \midrule
\multirow{1}{*}{1}	&	\multirow{1}{*}{\hypertarget{app:qid_19}{19}}	&	NL	&	How many capital cities does Australia have? \\ \cmidrule{3-4}
&	&	TBL	&	``Country ( exonym )'', ``Capital ( exonym )'', ``Country ( endonym )'', ``Capital ( endonym )'', ``Official or native language(s) (alphabet/script)'' \\ \cmidrule{3-4}
&	&	SQL (T)	&	{\ttfamily SELECT} count(Capital ( endonym )) {\ttfamily FROM} 1-1008653-1 {\ttfamily WHERE} Country ( endonym ) $=$ Australia \\ \cmidrule{3-4}
&	&	SQL (P)	&	{\ttfamily SELECT} count(Capital ( exonym )) {\ttfamily FROM} 1-1008653-1 {\ttfamily WHERE} Country ( exonym ) $=$ australia \\ \cmidrule{3-4}
&	&	ANS (T)	&	1 \\ \cmidrule{3-4}
&	&	ANS (P)	&	1 \\
\bottomrule
\end{longtable}

\end{document}